\documentclass[10pt,twocolumn,letterpaper]{article}

\usepackage{iccv}
\usepackage{times}
\usepackage{epsfig}
\usepackage{graphicx}
\usepackage{amsmath}
\usepackage{amssymb}
\usepackage{xcolor}
\usepackage{color}
\usepackage{times}
\usepackage{latexsym}
\usepackage{graphicx}
\usepackage{xcolor}
\usepackage{epsfig}
\usepackage{pxrubrica}        
\usepackage{url}
\usepackage{multirow}
\usepackage{booktabs}
\usepackage{cite}
\usepackage{subfigure}
\usepackage{amssymb}
\usepackage{here}
\usepackage{mathtools}
\usepackage{inconsolata}
\usepackage{microtype}

\usepackage[breaklinks=true,bookmarks=false]{hyperref}

\iccvfinalcopy 

\def\iccvPaperID{15} 

\ificcvfinal\fi

\begin{document}

\title{Pointing out Human Answer Mistakes in a Goal-Oriented Visual Dialogue}

\author{
    \text{Ryosuke Oshima}${}^1$\quad
    \text{Seitaro Shinagawa}${}^2$\quad
    \text{Hideki Tsunashima}${}^1$\quad
    \text{Qi Feng}${}^1$\quad
    \text{Shigeo Morishima}${}^3$\\
    ${}^1$Waseda University\quad
    ${}^2$Nara Institute of Science and Technology\\
    ${}^3$Waseda Research Institute for Science and Engineering\\
    {\tt\small \{ryosukeoshima@fuji, h.tsunashima@asagi, fengqi@ruri\}.waseda.jp, sei.shinagawa@is.naist.jp,  shigeo@waseda.jp}
}

\maketitle
\ificcvfinal\thispagestyle{empty}\fi

\begin{abstract}
   Effective communication between humans and intelligent agents has promising applications for solving complex problems. One such approach is visual dialogue, which leverages multimodal context to assist humans. However, real-world scenarios occasionally involve human mistakes, which can cause intelligent agents to fail. 
   While most prior research assumes perfect answers from human interlocutors, we focus on a setting where the agent points out unintentional mistakes for the interlocutor to review, better reflecting real-world situations.
   In this paper, we show that human answer mistakes depend on question type and QA turn in the visual dialogue by analyzing a previously unused data collection of human mistakes. We demonstrate the effectiveness of those factors for the model's accuracy in a pointing-human-mistake task through experiments using a simple MLP model and a Visual Language Model.
\end{abstract}

\section{Introduction}
\label{sec:Intro}
The communication between humans and intelligent agents has gained increasing attention due to its potential to solve various problems, such as making reservations and navigation.
To further enhance this capability, visual dialogue, which utilizes multimodal context, has emerged as a promising approach to assist humans~\cite{survey_visdialog}. 

In real-world scenarios, human interlocutors may not always respond accurately to agents due to misinterpretations or unintentional mistakes. In fact, human-to-human visual dialogue data collection includes failed tasks with incorrect human answers.
However, many current visual dialogue systems assume that the responses provided by the interlocutor are always correct~\cite{VilBERT, Uniqer}. 

To address the potential impact of human mistakes on visual dialogue accuracy, we introduce a task where an agent identifies and points out mistakes made by a human interlocutor in response to the agent's questions.
This task draws inspiration from Guess What?!~\cite{guesswhat}, a well-established benchmark for evaluating the performance of agents in visual dialogue tasks. 
We extend the original task by adding steps for the agent to point out mistakes to the human, who can then acknowledge and review their errors.

Previous work \cite{Non-Cooperative} examines situations where the interlocutor intentionally provides incorrect answers, emphasizing uncooperative behavior. However, encountering uncooperative dialogue partners is rare in typical problem-solving scenarios. In our study, we initiate goal-oriented dialogues with the assumption that the interlocutor is cooperative, providing an opportunity to resolve the task. 
Unlike \cite{Non-Cooperative}, our focus is on a scenario where the agent points out incorrect answers unintentionally provided by the interlocutor, which better reflects the real-world applications of visual dialogue.

\begin{figure}[t]
\setlength{\belowcaptionskip}{-0.6cm}
 \begin{center}
  \includegraphics[width=0.49\textwidth]{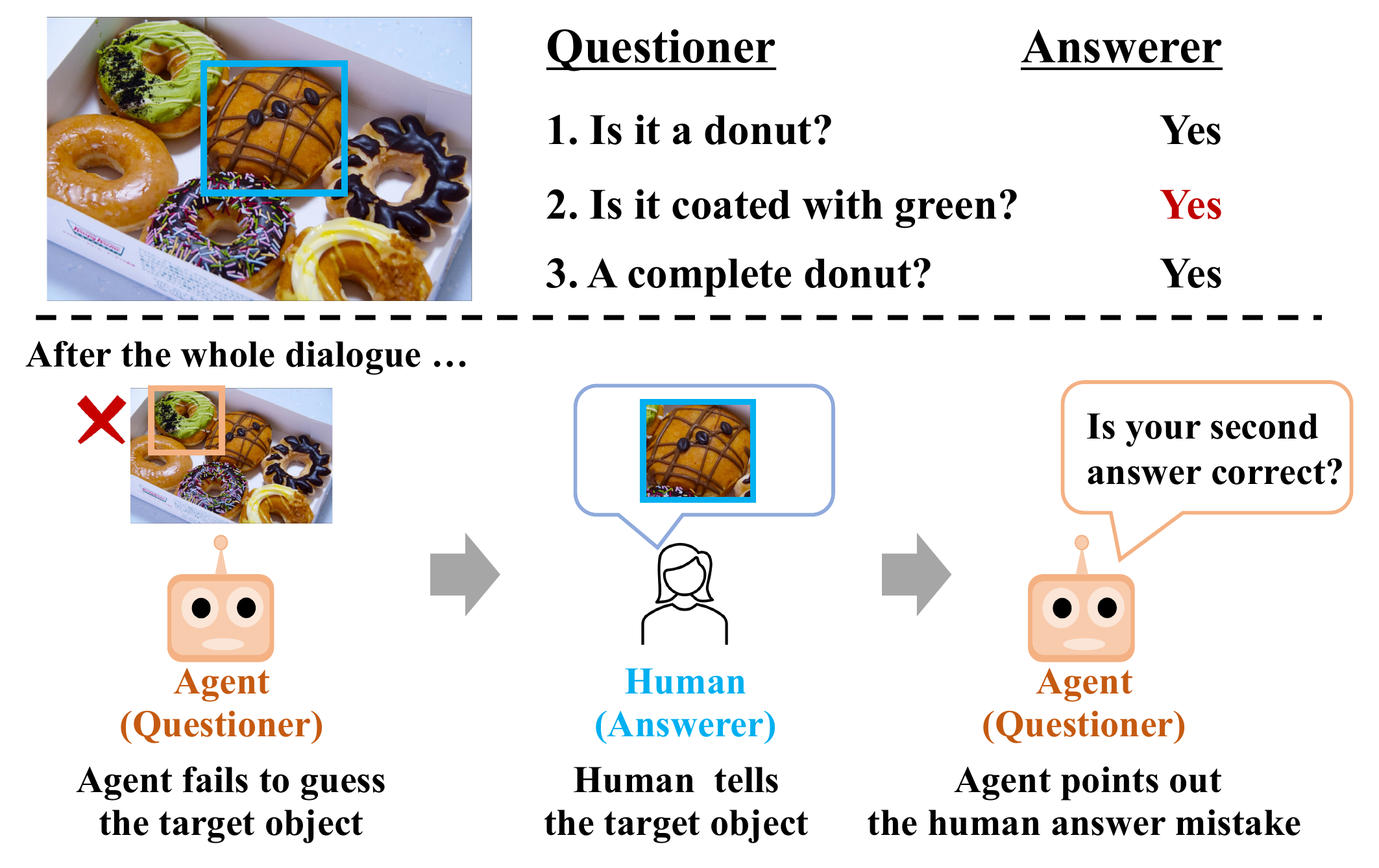}
  \caption{Overview of the pointing-human-mistake task. When the human interlocutor (Answerer) provides an incorrect answer ({\color[HTML]{FF0000}\textbf{red}}) about the target object ({\color[HTML] {3531FF}\textbf{blue}}) in a given image, the agent (Questioner) identifies and flags the mistake as a classification task. This process is illustrated in the bottom image.}
  \label{fig:guesswhat}
 \end{center}
\end{figure}
The problem setting of pointing out mistakes shares similarities with the problem setting of clarification requests for uncertain answers during dialogue~\cite{benotti-blackburn-2021-recipe, CR_CoD}. 
This approach proves valuable as it allows an agent to ask questions about ambiguous human statements, thereby mitigating the risk of failure. 
However, it is important to note that even with this questioning strategy, complete mistake prevention is not guaranteed, as other factors may contribute to task failures.
The problem setting of pointing out mistakes can be positioned as a problem of how to recover after a mistake has occurred and the task has failed, and this study aims to analyze the problem setting for this purpose.
In this paper, we focus on a pointing-human-mistake task, distinct from traditional clarification requests, which allows us to explore the potential of agents in identifying and pointing out human answer mistakes. Leveraging the widely used Guess What?! Dataset [5], we construct Human Mistake Dataset to facilitate our investigation.
Through a two-fold analysis, we establish clear correlations between the two key features and the occurrence of human mistakes: QA turns and Question types. Having identified the factors, we extend our study and implement them in a simple MLP model and a Visual Language Model (VLM). Through experiments, we demonstrate that incorporating these features leads to remarkable improvements in the agent's prediction performance, specifically in pointing out human mistakes during visual dialogues.

By providing insights into the importance of identifying human answer mistakes for enhancing multimodal communication in visual dialogue systems, our work offers valuable information for more effective and accurate visual dialogues in future applications.
\vspace{-2mm}
\section{Human Answer Mistake Analysis}
\subsection{The Pointing-human-mistakes Task} 
Guess What?! (Figure~\ref{fig:guesswhat} top) is a two-player game in which a Questioner asks yes/no questions to identify a target object, and an Answerer\footnote{\cite{guesswhat} calls an interlocutor Oracle. Instead, we use Answerer in this paper because an interlocutor can make mistakes and not give a perfect answer.} provides answers to those questions. 
In the pointing out human mistake task, an agent takes on the role of the Questioner, and a human interlocutor acts as the Answerer. 
The agent's goal is to point out mistakes made by the human interlocutor during the dialogue. 
Figure~\ref{fig:guesswhat} bottom provides an overview of the pointing-human-mistake task. 
By using this task, we can investigate an agent's ability to recognize and address human mistakes in a multi-turn conversational setting accurately.
\vspace{-1mm}
\subsection{Human Mistake Dataset Construction} 
\label{HED-construction}
To analyze the features of human answer mistakes and facilitate learning-based methods, we constructed a dataset based on Guess What?! \cite{guesswhat}, which consists of human-to-human dialogues including three types of collections: \emph{Success}, \emph{Failure}, and \emph{Incomplete}.
We bring attention to the \emph{Failure} collection that has never been used in ordinary visual dialogue settings. We constructed Human Mistake Dataset by randomly selecting 2,300 dialogues from the \emph{Failure} collection, filtering out noisy samples with small target object sizes, following \cite{dong-etal-2021-visually}.
As a result, Human Mistake Dataset contains 365 dialogues, enabling us to analyze human answer mistakes in visual dialogue settings.

\subsection{Analysis Methodology}
\label{subsec:analyze_mistake}
Firstly, we examine the relationship between the QA turn and the occurrence of answer mistakes. 
We hypothesize that answer mistakes are more likely to occur as the dialogue progresses, since subsequent turns tend to involve more complex objects or relations to determine the target object. 
Secondly, we explore the relationship between the question type and the occurrence of answer mistakes. 
This aspect is particularly relevant given the findings of \cite{they-not-alike} and \cite{beyond-task}, who reported that the correct answer rate of Answerer models varies depending on the question type. 
By analyzing the relationship between question type and answer mistakes, we aim to gain insights into the nature of human answer mistakes in visual dialogue settings.
\subsection{Analysis Results}
\label{subsec:analysis_results}
\paragraph{QA Turn}
\begin{figure}[t]
\setlength{\belowcaptionskip}{-0.8cm}
\begin{center}
   \includegraphics[width=0.49\textwidth]{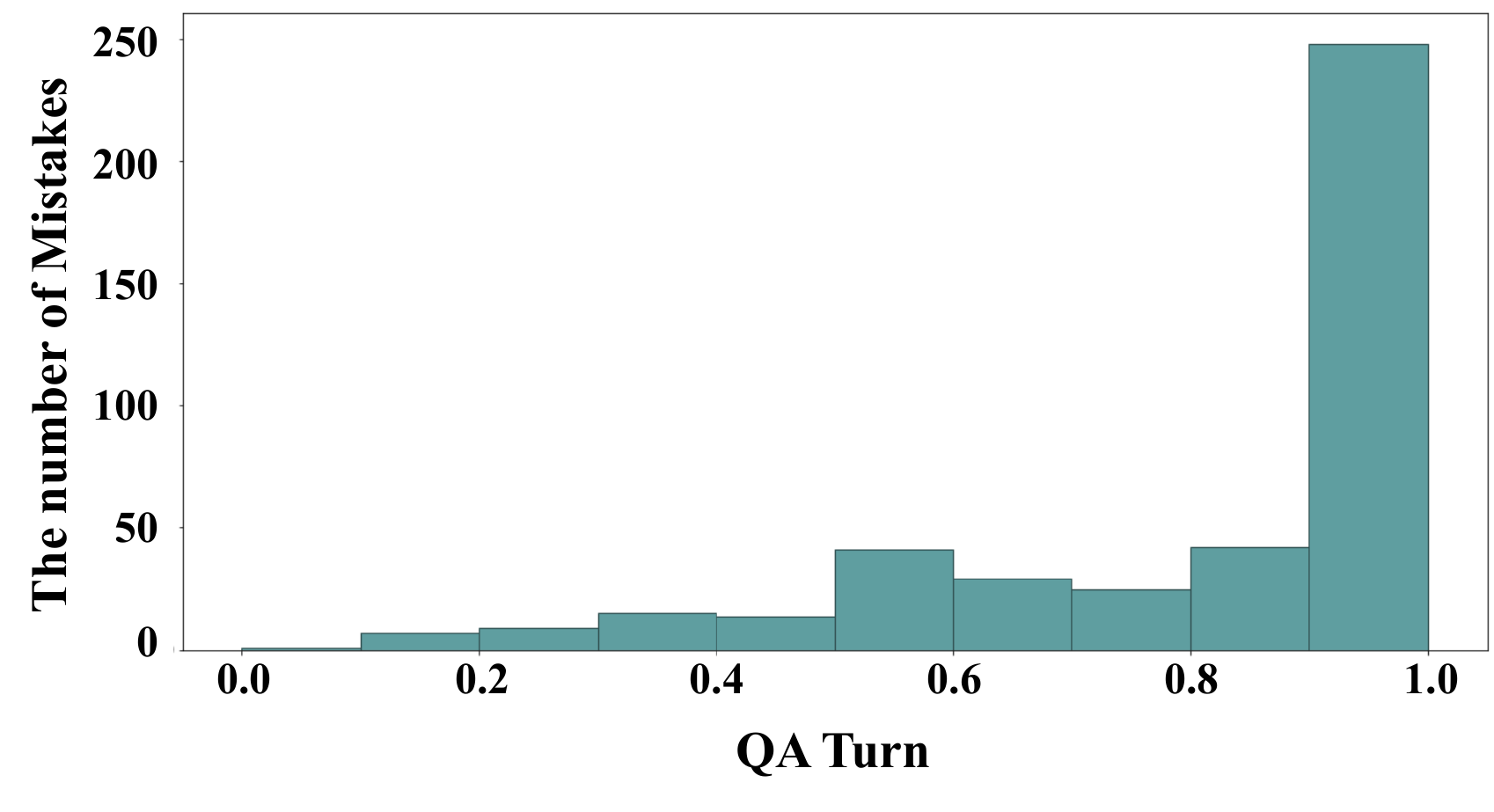}
   \vspace{-2mm}
   \caption{Histogram of relationship with QA turn and answer mistakes. The horizontal axis is normalized.}
   \label{fig:mistake_pos}
 \end{center}
\end{figure}
\vspace{-2mm}
\begin{figure}[t]
\setlength{\belowcaptionskip}{-0.8cm}
\begin{center}
\includegraphics[width=0.49\textwidth]{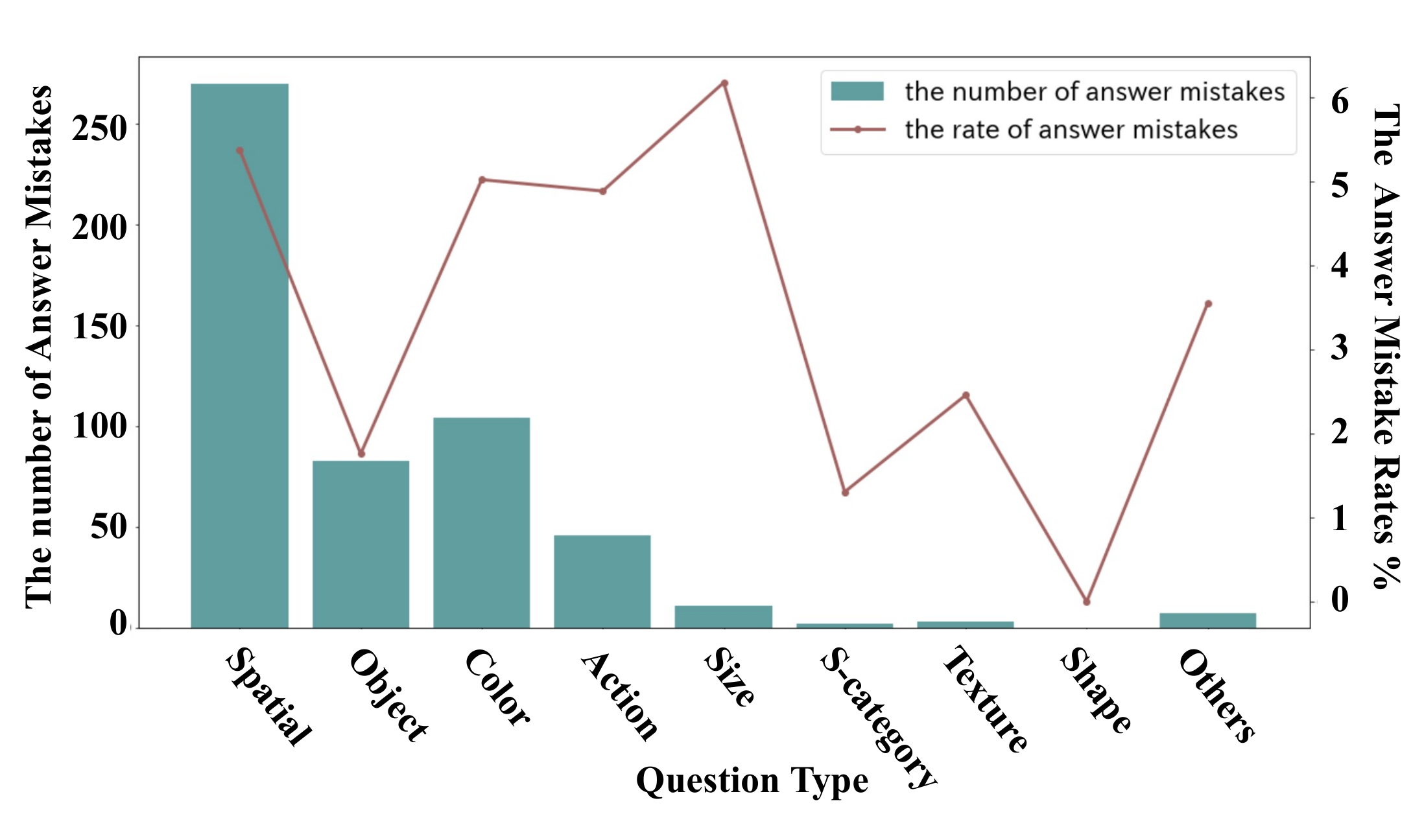}
\vspace{-5mm}
   \caption{The mistake question rate. S-category means Super-category type. It is essential to note that the rates represent the mistake rates for each question type; therefore, the sum of all the rates does not equal 100\%.}
   \label{fig:q-type}
 \end{center}
 \vspace{-6mm}
\end{figure}
Figure~\ref{fig:mistake_pos} illustrates the relationship between the number of answer mistakes and the turn number in the dialogues. 
The QA turn (horizontal axis) is normalized as $\frac{\text{current turn}}{\text{total turns}}$, which represents a relative turn in the dialogue because the number of turns varies in dialogues. 
As expected, the frequency of answer mistakes increases as the dialogue progresses, particularly in the latter half of the dialogue. 
Notably, 231 out of 431 samples of answer mistakes occurred in the last turn. (See Appendix~\ref{appendix:qa_turn_analysis} for details)

\vspace{-3mm}
\paragraph{Question Type}
We investigated the impact of question types on human answer mistakes. 
We hypothesized that answer mistakes vary by the question types and conducted Fisher's exact test (refer to Appendix \ref{appendix:question_type} and \ref{appendix:q_type_test} for question types and details).
Our findings showed that at least one question type has a significantly higher mistake rate than others, which supports our hypothesis.

Figure \ref{fig:q-type} illustrates the answer mistake rates for each question type. 
We calculated the mistake rates by dividing the number of incorrect answers by the total number of answers for each question type. 
The results indicate that humans are more likely to make mistakes when answering Spatial, Color, Action, and Size questions
(See Appendix \ref{appendix:q_type_test} for details.
\vspace{-3mm}
\section{Model}
We employ two types of models and observe how effectively our findings in Section \ref{subsec:analysis_results} improve the accuracy of the mistake-pointing out model.
\vspace{-2mm}
\subsection{MLP Model}
\label{Model:mlp_model}
\begin{figure}[t]
\setlength{\belowcaptionskip}{-0.6cm}

\begin{center}
    \includegraphics[width=0.49\textwidth]{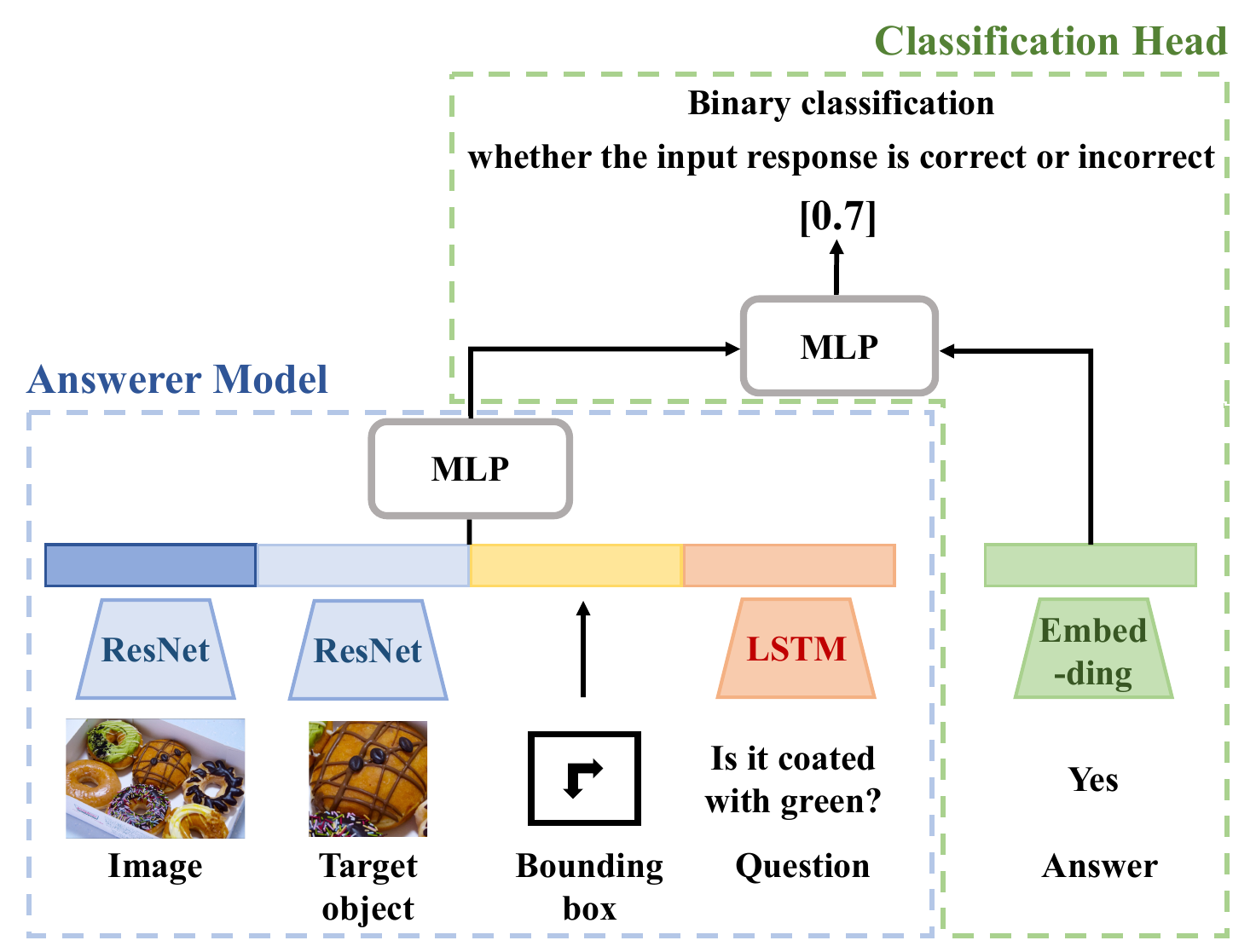}
   \caption{Baseline model overview diagram}
   \label{fig:model}
 \end{center}
\vspace{-2mm}
\end{figure}

\paragraph{Model Details}
We extend the Answerer model proposed in~\cite{guesswhat} by adding a classification head for detecting mistakes in human answers (Figure \ref{fig:model}).
We set a threshold of 0.5 for the model's output, allowing it to perform binary classification on whether the answer is {correct or incorrect}. 
We call this model the baseline model.
While there are many possible methods to incorporate Question type and QA turn into the baseline model, we opt for a straightforward approach of introducing a new input to the baseline model (details provided in Appendix \ref{appendix:models}).
We call these models the Question type model and the QA turn model, respectively.
\vspace{-5mm}
\paragraph{Pretraining Strategy}
Human mistakes in answers during dialogue are rare, making it costly to collect many human answer mistakes as it requires reviewing the entire dialogue and finding mistakes afterward. 
To overcome this challenge, we introduce pretraining with Synthetic Dataset using the \emph{Success} collection of Guess What?! \cite{guesswhat}. 
Specifically, we randomly flipped Yes/No in human answers and built a dataset of approximately 131k dialogues.
Preliminary experiments showed that this pretraining strategy was effective (details are shown in Appendix \ref{appendix:pretrainig}).
To prevent the model from learning an unnatural dialogue flow in Synthetic Dataset (details are shown in Appendix \ref{appendix:automatic_dialog}), we did not deliberately include the dialogue history as an input to the model in this study.
\vspace{-4mm}
\paragraph{Training Details}
For pre-training, we use 70\% of the Synthetic Dataset for training and 15\% for validation, similar to~\cite{guesswhat}. We omit the remaining 15\% of the test dataset as the models are not evaluated during pre-training.
We perform k-fold cross-validation with $k=4$ while fine-tuning on Human Mistake Dataset. The results in Section~\ref{subsec:mlp_model} represent the average results for each experiment across the k cross-validation iterations.
We set the learning rate to $10^{-4}$ for pre-training and $10^{-5}$ for fine-tuning with the Adam~\cite{ADAM} optimizer.

\subsection{Visual Language Model (VLM)}
We further explore VLMs integrated with large language models, which have recently achieved high performance without task-specific fine-tuning~\cite{MME}.
We examine a representative VLM, OpenFlamingo\footnote{\url{https://github.com/mlfoundations/open_flamingo}}~\cite{NEURIPS2022_960a172b}, in in-context learning setting to investigate VLMs' ability to point out human mistakes with prompts. Despite testing other VLM models (InstructBLIP~\cite{InstructBLIP} and BLIP2~\cite{pmlr-v202-li23q}), OpenFlamingo outperformed them in our experiment (see Appendix \ref{appendix:vlm_exp_details} for details).

\vspace{-3mm}
\section{Experiment}
In this study, we conduct an experiment to explore how incorporating the mistake tendency identified in Section \ref{subsec:analysis_results} can improve the models' accuracy.
\vspace{-4mm}
\paragraph{Dataset}
Our study uses two datasets: Synthetic Dataset for pre-training and Human Mistake Dataset for fine-tuning and evaluation. Human Mistake Dataset is divided into two parts: the same image dataset using the same images as the training data and the different image dataset with different images. It is noticed that both datasets are unknown to the dialogue text.
We use both the same and different image datasets for evaluating MLP models and the different image dataset for evaluating a VLM (see Appendix \ref{appendix:how-to-use-dataset} in detail).
\vspace{-3mm}
\paragraph{Evaluation}
To evaluate the performance of the model in pointing out answer mistakes, we use the F-score, Recall, and Precision which can evaluate imbalanced data (see Appendix \ref{appendix:f-scores}). 
Human Mistake Dataset used for evaluation is imbalanced; the number of positive cases (answer mistakes) is smaller than that of negative cases (correct answers). 
This is due to the rare occurrence of answer mistakes.
\vspace{-2mm}
\section{Results}
\label{sec:results}
\subsection{MLP Models}
\label{subsec:mlp_model}
We tested the baseline model, QA turn model, and Question type model to see if these additional features could improve the model's ability to point out mistakes in human answers based on the insights gained from our analysis in Section \ref{subsec:analysis_results}.
We oversampled\footnote{A term used in the field of imbalanced data, and refers to the process of increasing the number of samples in categories with fewer instances.} samples in the fine-tuning stage to ensure an equal number of samples for each question type. Table \ref{tab:add_input} shows the results of MLP models' experiments.

The baseline model performed best in the same image dataset, whereas the QA turn model and Question type model outperformed it in the different image dataset. 
However, we were concerned that the accuracy of pointing out mistakes might decrease for responses other than the last one.
Therefore, we evaluated the accuracy of the QA turn model both at the last time and at all other times. 
Table \ref{tab:last_no_last_result} shows that the QA turn model improved the F-score for the last response in the different image dataset without a significant decrease for other responses.
The results indicate that in the {\em same} image dataset, including the additional input from the QA turn and Question type models led to unnecessary complexity, resulting in a lower F-score than the baseline model, which already captures visual information in known images. 
Conversely, in the {\em different} image dataset, the baseline model was less effective in capturing visual information, making the input from the QA turn and Question type models valuable. Furthermore, Table \ref{tab:last_no_last_result} demonstrates that the QA turn model may serve as a useful predictor instead of solely relying on learning the data distribution.
\vspace{-1mm}
\subsection{Visual Language Model}
\label{results:foundation_model}
To assess the effectiveness of the human mistake features identified in Section \ref{subsec:analysis_results}, we also conducted tests using a VLM, which has recently been actively studied for its high accuracy. In particular, we used OpenFlamingo~\cite{NEURIPS2022_960a172b} and provided three types of prompts as input: (1) a normal prompt to point out human answer mistakes; (2) a question type hint prompt that gives information that humans are more likely to make mistakes in their responses to some question types; (3) a QA turn hint prompt to provide information that humans tend to make more mistakes as the dialogue progresses.
We only conducted experiments with the different image dataset for OpenFlamingo.
We also experimented with the case in which OpenFlamingo got a dialogue history. 
The reasons for these choices are provided in Appendix \ref{appendix:vlm_exp_details}.
It is worth noting that in-context learning is known to be highly sensitive
to few-shot prompting~\cite{pmlr-v139-zhao21c, NEURIPS2021_5c049256}, and we have not been able to exhaustively examine all possible prompts.


Table \ref{tab:foundation_model_add_input_history} shows the results with OpenFlamingo~\cite{NEURIPS2022_960a172b}.
OpenFlamingo's F-score is lower than 38\%, worse than the MLP models. It indicates that pointing out human answer mistakes seems difficult for OpenFlamingo without task-specific fine-tuning, unlike other typical V\&L tasks such as VQA.
We found that OpenFlamingo with Question type and QA Turn improved prediction accuracy. Human features are also effective for OpenFlamingo, whether or not dialogue history is considered.
\begin{table}[t]
\centering
\small
\begin{tabular}{ccc}
\hline
Model                & Same image     & Different image                               \\ \hline
Baseline         & \textbf{0.811} & 0.482                                 \\ 
QA turn   & 0.718                                 & {\color[HTML]{3531FF}\textbf{0.514}} \\ 

Question type & {\color[HTML]{3531FF}\textbf{0.743}}                        &  \textbf{0.527} \\ \hline
\end{tabular}
\vspace{1mm}
\caption{The results of MLP models. From the top, the results are for the baseline, the QA turn model, and the Question type model. The score is F-score. The best score is in \textbf{black bold}, and the second-best score is in {\color[HTML] {3531FF}\textbf{blue}}.}
\label{tab:add_input}
\end{table}
\begin{table}[t!]
\setlength{\belowcaptionskip}{-0.6cm}
\centering
\small
\begin{tabular}{cccccc}
\hline
\multirow{2}{*}{Model} & \multicolumn{2}{c}{Last time}                        &  & \multicolumn{2}{c}{Other than last time} \\ \cline{2-3} \cline{5-6} 
                                  & Same                          & Different                        &  & Same                            & Different  \\ \cline{1-3} \cline{5-6} 
Baseline                              & \textbf{0.875} & 0.548                         &  & \textbf{0.714}   & 0.406   \\
QA turn             & 0.789                         & \textbf{0.608} &  & 0.609                           & 0.406   \\ \hline
\end{tabular}
\vspace{1mm}
\caption{F-scores at the last time and at other times for the baseline model and QA turn model. Same and Different means the same image dataset and the different image dataset, respectively.}
\label{tab:last_no_last_result}
\end{table}

\begin{table}[]
\centering
\small
\begin{tabular}{ccc}
\hline
Prompt type        & Without History & With history \\ \hline
Normal             &     0.313     &     0.325    \\
QA turn hint       & \textbf{0.374}     &    \textbf{0.377}  \\
Question type hint & {\color[HTML]{3531FF}\textbf{0.366}}     &  {\color[HTML]{3531FF}\textbf{0.372}}  \\ \hline
\end{tabular}
\vspace{1mm}
\caption{The results of OpenFlamingo. The top three lines are the result of not including the dialogue history in the prompt, and the bottom three lines are the result of including the dialogue history in the prompt.}
\label{tab:foundation_model_add_input_history}
\end{table}
\vspace{-2mm}
\section{Conclusion}
\vspace{-1mm}
In this paper, We considered a task in which agents point out human answer mistakes and analyzed what would be the key factors of human answer mistakes. 
We observed that mistakes were more common towards the end of the dialogue and varied based on question types.
By incorporating these human mistake features, we enhanced the performance of both the MLP model and the VLM in our experiments on the actual pointing-human-mistake task.
In future work, we aim to explore the generalizability of our findings to other goal-oriented visual dialogues.
\vspace{-2mm}
\paragraph{Acknowledgements}
{\footnotesize
This work was supported by JSPS KAKENHI Grant Numbers JP21H05054, JP21K17806.
}

\clearpage
\appendix
\section{Question Type}
\label{appendix:question_type}
\begin{table}[h]
\centering
\begin{tabular}{c|c}
\hline
Question type  & Example sentence \\ \hline
Spatial        & On the right side half?                \\
Object         & Is it a car?                \\
Color          & Is it white?                \\
Action         & Are they wearing jeans?              \\
Size           & A small one?                \\
Super-category & Is the object electronic?              \\
Texture        & Is it made of metal?                \\
Shape          & Is it a round container?                \\
Others         & Is it edible?                \\ \hline
\end{tabular}
\vspace{1mm}
\caption{Example sentences of each question type.}
\label{tab:question_example}
\end{table}
Table \ref{tab:question_example} shows example sentences for each question type.
Question types are labeled using the keyword matching method, categorizing question types using keywords like {\em left} for the Spatial category.
Super-category is a higher-level group containing related sub-categories, organizing information hierarchically, such as fruit for banana or vehicle for car.
\section{Question Types Analysis}
\label{appendix:q_type_test}
\paragraph{Hypothesis Testing}
Our hypothesis is that {\em the rate of human answer mistakes varies by question type}. 
Therefore, if we can establish that any question type is significantly more likely to be incorrect than another, it would support our hypothesis. 
Since we are focusing on the difference in the proportion of mistakes in each question type, we conducted hypothesis testing for the difference in the Population proportions.

Moreover, it's important to note that not all answers are provided by the same person; instead, crowd workers contribute answers. As we are dealing with 9 question types, this constitutes an independent test involving more than two groups. As a result, we conducted Fisher's exact rate test.
We set the significance level at 1$\%$ and performed an upper one-tailed test. The resulting p-value is $0.0004998 (< 0.01)$, which supports our hypothesis that human answer mistakes vary by question type.
\paragraph{Trend of mistaken Question Types}
Humans are more likely to make mistakes when answering Spatial, Color, Action, and Size questions. 
One possible reason is that these question types are more difficult than the other question types (Shape, Texture, Object, and Super-category).
Most Shape and Texture questions are easy-to-understand, such as ``round,'' ``square,'' ``wooden,'' and ``metal.''
Most Object and Super-category questions are monotonous, such as `Is it a banana?' Answerer could understand the meaning of the questions by reading a single word without considering it as a question sentence.
In contrast, the mistake rates for Action and Spatial questions were higher because Answerer needs to take these questions as sentences, which means they are more difficult to understand.
\section{QA Turn Analysis}
\label{appendix:qa_turn_analysis}
231 out of 431 samples of answer mistakes occurred in the last turn.
This trend is possibly related to a bias on the problem setting.
While a mistake in a dialogue would be pointed out instantly, a mistake in the last turn has no chance to be recovered.
The former mistake cases would be included {\em Success}, not {\em Failure} collection used in our Human Mistake Dataset.
\section{Pre-training with Synthetic Dataset}
\label{appendix:pretrainig}
\begin{table}[h]
\centering
\small
\begin{tabular}{ccc}
\hline
Learning method       & Same image                            & Different image                          \\ \hline
Human mistake                  & 0.730                           & 0.368                                             \\
Synthetic + Human mistake      & \textbf{0.811} & \textbf{0.482} \\ \hline
\end{tabular}
\caption{The results of each learning method. The score is F-score. Same image and different image mean the results of the same image dataset and the different image dataset, respectively.}
\label{tab:compare_learning_method}
\end{table}
We conducted the experiment to investigate the effectiveness of pre-training with Synthetic Dataset in addressing the challenge of limited erroneous human answers in training data. 
Specifically, we compared two learning methods: (1) training the model solely with Human Mistake Dataset, and (2) pre-training the model with Synthetic Dataset and fine-tuning it using Human Mistake Dataset. Due to fewer positive examples (i.e., human answer mistakes) than negative examples (i.e., correct answers), we oversampled the training data during the fine-tuning process.

Table \ref{tab:compare_learning_method} presents the experiment's results. The pre-training with Synthetic Dataset in this study achieved the highest F-score. 
This indicates that the pre-training strategy, which employs the Synthetic Dataset even in the absence of human mistakes, is effective.
\section{Unnatural Dialogue Flow}
\label{appendix:automatic_dialog}
\begin{figure}[h]
 \begin{center}
   \includegraphics[width=0.50\textwidth]{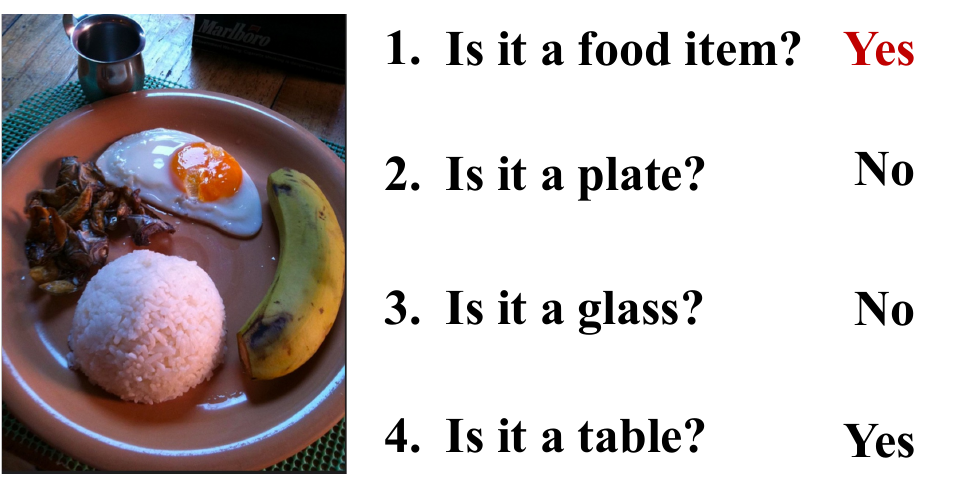}
   \caption{An example of unnatural dialogue flow in Synthetic Dataset, where the first answer is incorrect.}
   \label{fig:automatic_dialog}
 \end{center}
\end{figure} 
Figure \ref{fig:automatic_dialog} shows an example of unnatural dialogue flow. 
Although it is confirmed that the topic is food at time 1, the question `{\em Is it a plate?}' is asked at time 2. 
In a natural dialogue, Answerer is expected to ask questions about food, such as {\em Is it a banana?}.
\section{Evaluation for Imbalanced Data}
\label{appendix:f-scores}
Accuracy $\left(=\frac{\mbox{correct answers}}{\mbox{sample size}}\right)$ is not an appropriate evaluation metric in imbalanced data.
The reason is that if the model learns to predict negative cases with many samples, the model can achieve high accuracy, even though it cannot predict positive cases at all.
Instead, we use the F-score defined by Recall and Precision. It serves as one of the evaluation metrics for imbalanced data.
The true positive (correctly predicting a positive label as positive), true negative (correctly predicting a negative label as negative), false positive (incorrectly predicting a negative label as positive), and false negative (incorrectly predicting a positive label as negative) can be used to calculate Recall and Precision as follows:
\begin{equation}
    \mbox{Recall} = \frac{\mbox{True Positives}}{\mbox{True Positives}+\mbox{False Negatives}}
\end{equation}
\begin{equation}
    \mbox{Precision} = \frac{\mbox{True Postives}}{\mbox{True Postives}+\mbox{False Postives}}
\end{equation}
In imbalanced data, if a model simply learns to predict many instances as "positive," the Recall will be high while the Precision will be low. Conversely, if the model learns to predict many instances as "negative," the Precision will be high while the Recall will be low.

The F-score is used to appropriately evaluate such models, as it represents the harmonic mean of Recall and Precision. This metric considers both Recall and Precision, providing a more balanced assessment of the model's performance in an imbalanced dataset.

\begin{equation}
    \mbox{F-score} = 2\times\frac{\mbox{Precision}\times\mbox{Recall}}{\mbox{Precision}+\mbox{Recall}}
\end{equation}
\section{Model Details}
\label{appendix:models}
We describe below the details of the baseline model.
We get the embedding of the whole image $I_{emb}$ and that of the target object's cropped image $S_{emb}$ using ResNet~\cite{ResNet}.
We also get the embedding of the target object's spatial information $x_{spatial}$ from the bounding box, following \cite{guesswhat}.
\begin{multline}
    x_{spatial} = [x_{min}, y_{min}, x_{max}, y_{max}, \\ x_{center}, y_{center}, w_{box}, h_{box}]
\end{multline}
Next, $I_{emb}$, $S_{emb}$, $x_{spatial}$, and $q_t^{emb}$, the embedding of the question $q_t$ at time t encoded by LSTM~\cite{LSTM}, are concatenated and passed through MLP layers to generate a semantic vector $q_{mean}$ for each question.
\begin{equation}
\label{q_mean}
    q_{mean} = \textnormal{MLP}_m \left(\left[I_{emb}; S_{emb}; x;  q_t^{emb}\right]\right)
\end{equation}
$[\cdot ; \cdot]$ denotes vector concatenation.
We obtain the probability that the human interlocutor's answer is incorrect $p_m$:
\begin{equation}
\label{p_mistake}
p_{m} = \textnormal{sigmoid}\left(\textnormal{MLP}_c \left(\left[q_{mean}; a_t^{emb}\right]\right)\right)
\end{equation}
Finally, we determine whether an answer is correct or incorrect by a threshold value.
\begin{equation}
\label{judge_mistake}
\left\{
\begin{array}{ll}
1 \quad \mbox{(Incorrect answer)} & p_m > 0.5 \\
0 \quad \mbox{(Correct answer)} & p_m \leq 0.5
\end{array}
\right.
\end{equation}
We do not add the category label of the target object, such as a dog or banana, because the model relies on the category label and its spatial information instead of the visual features of the image as ~\cite{they-not-alike} mentioned.

\begin{figure}[H]
 \begin{center}
  \includegraphics[width=0.49\textwidth]{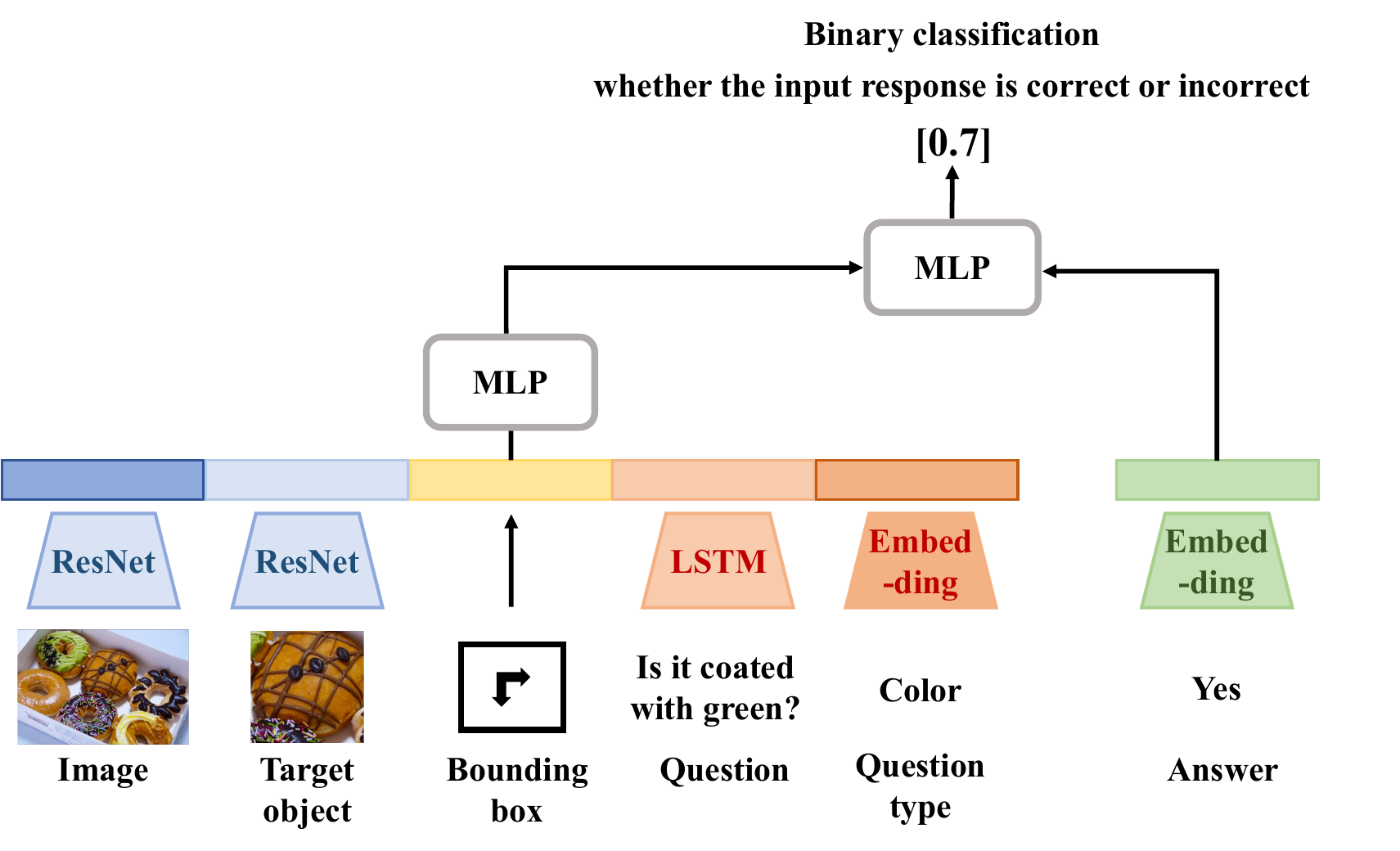}
  \caption{Question type model}
  \label{fig:model_qcat}                                                    
 \end{center}
\end{figure}

\begin{figure}[H]
 \begin{center}
  \includegraphics[width=0.49\textwidth]{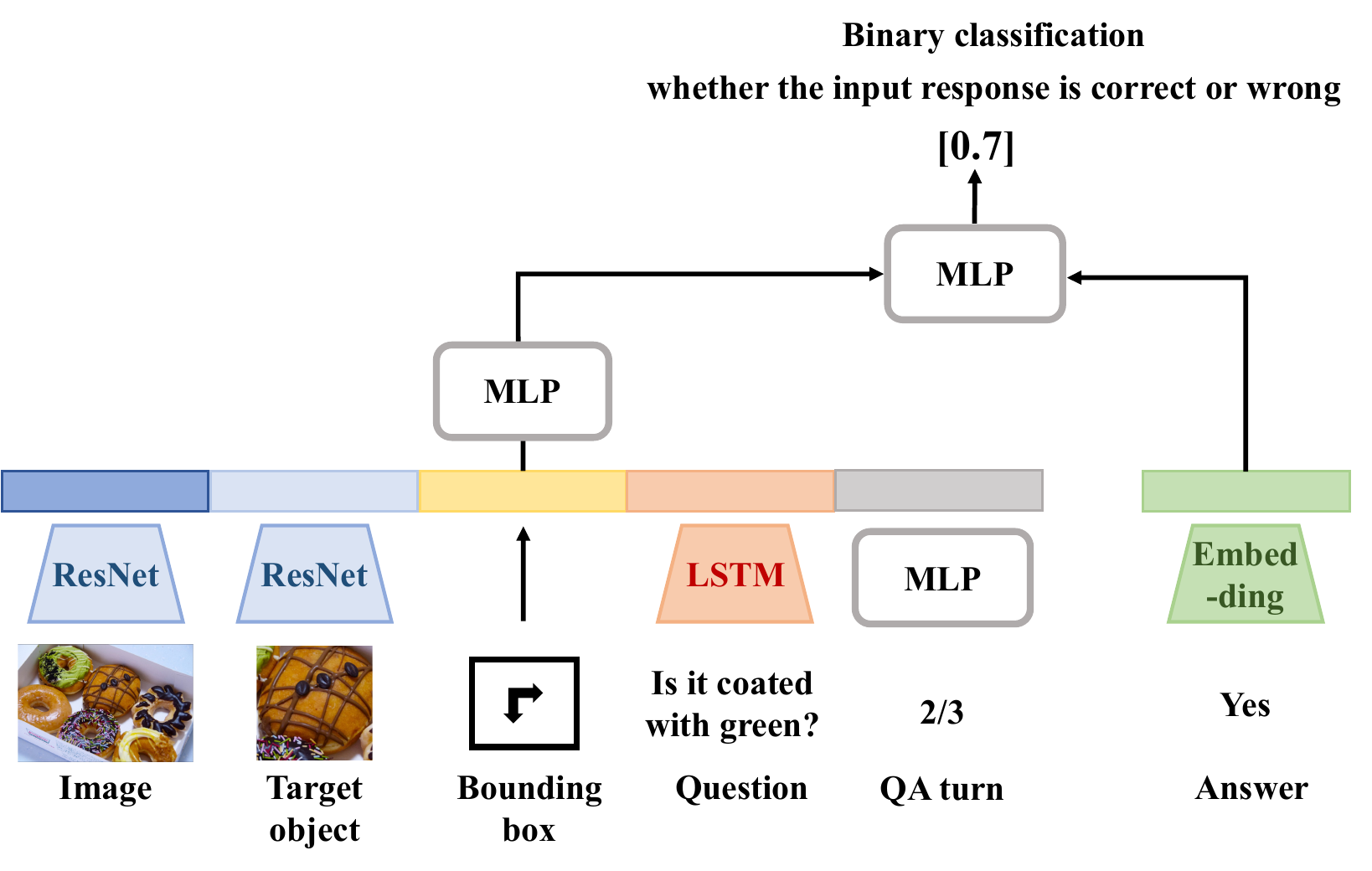}
  \caption{QA turn model}
  \label{fig:model_timelab}
 \end{center}
\end{figure}

Figure \ref{fig:model_qcat}, \ref{fig:model_timelab} shows the Question type model and QA turn model, respectively. 
We get the embedding of question type $q_{type}^{emb}$ and QA turn $a_{time}^{emb}$ by the embedding layer and the MLP layers, respectively.
We obtain $q_{mean}$ using equation (\ref{equation:q_type}) for the Question type model and equation (\ref{equation:q_time}) for the QA turn model.
\mathtoolsset{showonlyrefs=false}
\begin{equation}
\label{equation:q_type}
    q_{mean} = \textnormal{MLP}_m \left(\left[I_{emb}; S_{emb}; x;  q_t^{emb}; q_{type}^{emb}\right]\right)
\end{equation}
\begin{equation}
\label{equation:q_time}
    q_{mean} = \textnormal{MLP}_m \left(\left[I_{emb}; S_{emb}; x;  q_t^{emb}; q_{time}^{emb}\right]\right)
\end{equation}

\section{VLM Experiment Details}
\label{appendix:vlm_exp_details}
\begin{table*}[]
\centering
\begin{tabular}{ll}
\hline
Type         & Prompt                                                    \\ 
\hline
Normal       & \begin{tabular}[c]{@{}l@{}}
$<$BOS$>$ $<$image$>$ 
\\ The target object: $\{$position: a yellow rectangle, name: $\{$object category$\}$$\}$,\\ Question: $\{$question text$\}$, 
\\ Answer: $\{$answer text$\}$, 
\\ Judge: Is this answer a mistake? 
\\ Output: $\{$answer$\}$.$<$EOC$>$
\end{tabular} \\
\hline
Qtype        & \begin{tabular}[c]{@{}l@{}}
$<$BOS$>$ $<$image$>$\\ 
The target object: $\{$position: a yellow rectangle, name: $\{$object category$\}$$\}$,\\ 
Question: $\{$question text$\}$,\\
Answer: $\{$answer text$\}$,\\
This question type: $\{$qtype$\}$\\ 
Hint: $<$spatial$>$, $<$color$>$, $<$action$>$, and $<$size$>$ questions are easy to make mistakes on.\\
Judge: Is this answer a mistake?\\
Output: $\{$answer$\}$.$<$EOC$>$
\end{tabular} \\
\hline
Time         & \begin{tabular}[c]{@{}l@{}}
$<$BOS$>$ $<$image$>$\\ 
The target object: $\{$position: a yellow rectangle, name: $\{$object category$\}$$\}$,\\ 
 Question at $\{$answer time$\}$ progression of dialogue: $\{$question text$\}$,\\ 
Answer: $\{$answer text$\}$,\\
Hint: The frequency of answer errors increases as answer time is bigger.\\
Judge: Is this answer a mistake?\\
Output: $\{$answer$\}$.$<$EOC$>$
\end{tabular} \\ 
\hline
Normal (history) & \begin{tabular}[c]{@{}l@{}}$<$BOS$>$ $<$image$>$ 
\\ The target object: $\{$position: a yellow rectangle, name: $\{$object category$\}$$\}$,\\
Dialogue history: $\{$history$\}$,\\
Question: $\{$question text$\}$, \\ Answer: $\{$answer text$\}$, 
\\ Judge: Is this answer a mistake? 
\\ Output: $\{$answer$\}$.$<$EOC$>$
\end{tabular} \\
\hline
Qtype (history)  & \begin{tabular}[c]{@{}l@{}}
$<$BOS$>$ $<$image$>$\\ 
The target object: $\{$position: a yellow rectangle, name: $\{$object category$\}$$\}$,\\ 
Dialogue history: $\{$history$\}$,\\
Question: $\{$question text$\}$,\\
Answer: $\{$answer text$\}$,\\
This question type: $\{$qtype$\}$\\ 
Hint: $<$spatial$>$, $<$color$>$, $<$action$>$, and $<$size$>$ questions are easy to make mistakes on.\\
Judge: Is this answer a mistake?\\
Output: $\{$answer$\}$.$<$EOC$>$
\end{tabular} \\
\hline
Time (history)   & \begin{tabular}[c]{@{}l@{}}
$<$BOS$>$ $<$image$>$\\ 
The target object: $\{$position: a yellow rectangle, name: $\{$object category$\}$$\}$,\\ 
Dialogue history: $\{$history$\}$,\\
 Question at $\{$answer time$\}$ progression of dialogue: $\{$question text$\}$,\\ 
Answer: $\{$answer text$\}$,\\
Hint: The frequency of answer errors increases as answer time is bigger.\\
Judge: Is this answer a mistake?\\
Output: $\{$answer$\}$.$<$EOC$>$\end{tabular} \\ \hline
\end{tabular}
\vspace{2mm}
\label{tab:open_flamingo_prompt}
\caption{Examples of prompts for each type. $<$image$>$ takes as input the embedding of the image. $\{$object category$\}$ is the object category name of each target object (e.g., donut, vase), $\{$question text$\}$ and $\{$answer text$\}$ is the question and answer (yes or no) of the corresponding part to judge whether it is a mistake or not, and $\{$answer$\}$ is the result of judging whether the corresponding response is a mistake or not. $\{$qytpe$\}$ contains the question type (e.g., $<$color$>$) and $\{$answer time$\}$ contains the value of $\frac{\text{current turn}}{\text{total turns}}$.}
\end{table*}

\begin{figure*}[h]
 \begin{center}
   \includegraphics[width=0.95\textwidth]{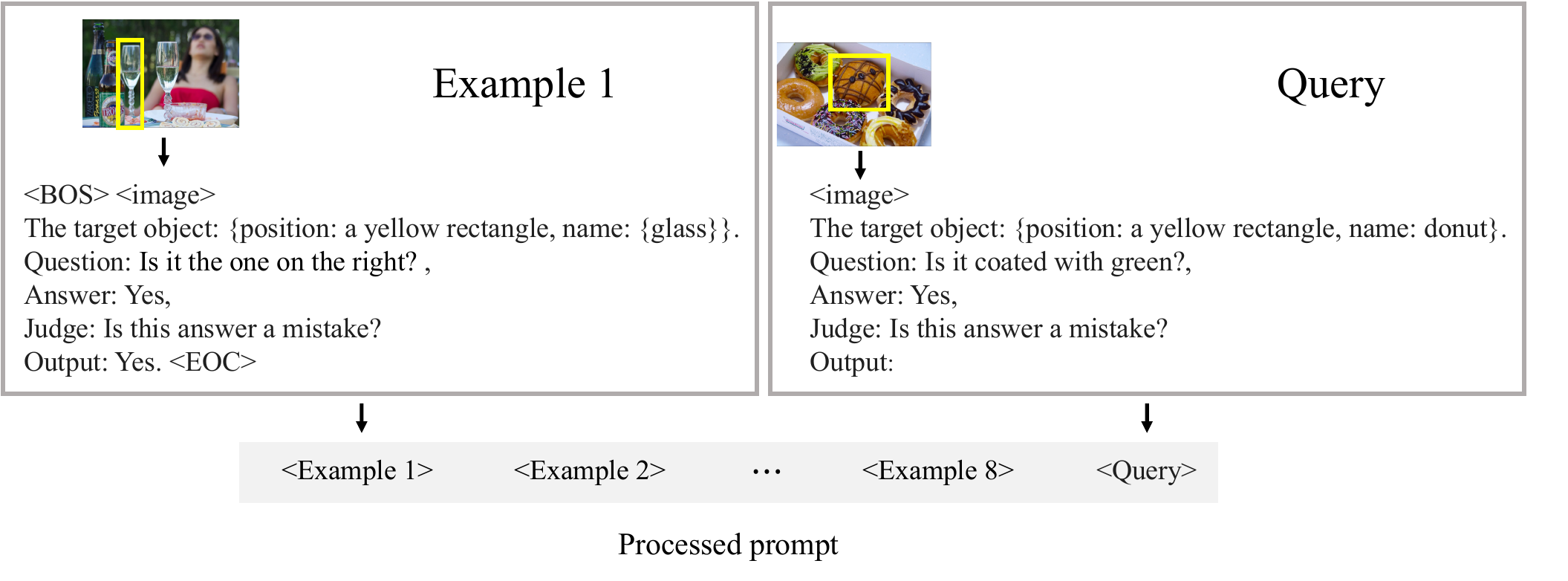}
   \caption{Few-shot prompt overview diagram. }
   \label{fig:fewshot}
 \end{center}
\end{figure*} 
\label{appendix:foudation_prompts}
\paragraph{Evaluation}
We only evaluated OpenFlamingo~\cite{NEURIPS2022_960a172b} using the different image dataset to ensure a fair comparison with the MLP model. This was necessary because OpenFlamingo was not specifically tuned with the exact same image dataset as the MLP model.
\paragraph{Inputs and Prompts}
We used examples of actual human mistakes for the few-shot prompting, rather than a sample from synthetically created mistakes. In particular, we randomly sampled eight examples from the same image dataset used for fine-tuning the MLP model.
We conducted preliminary experiments in an input format similar to MLP, where the object's crop image and the bounding box's position were given in list format, but the F-score was very low. We provided the target object information with the object's position in the image surrounded by a yellow bounding box and the object's name.

OpenFlamingo does not learn the unnatural dialogue flow by including the dialogue history, because it is not pre-trained with the Synthetic Dataset. We conducted experiments with both prompts with and without dialogue history.
Figure \ref{fig:fewshot} shows an overview of the eight few-shot prompts added as input to OpenFlamingo.
Table \ref{tab:open_flamingo_prompt} shows the prompts for each prompt type.
We provided the prompts in a structured JSON-like format.
\paragraph{Other VLM Models}
We also conducted experiments with Instruct BLIP~\cite{InstructBLIP} and BLIP2~\cite{pmlr-v202-li23q} as well as OpenFlamingo. 
However, with and without Instruction following, the F-score was very low, about 25\%.
We think this is because in-context learning did not work well, as BLIP2 was trained with the pre-training dataset, which only contains a single image-text pair per sample, as mentioned in ~\cite{pmlr-v202-li23q}.
OpenFlamingo trained with MMC4~\cite{zhu2023multimodal}, which  includes documents sourced from web scraping, interleaved images, and text, with multiple image-text combinations in each sequence.
\section{How to Use Each Dataset}
\label{appendix:how-to-use-dataset}
Table~\ref{tab:how-to-use-dataset-mlp} shows how we use Synthetic Dataset, the same image dataset, and the different image dataset when we conduct MLP models' experiments.
\begin{table*}[]
\centering
\begin{tabular}{cccccccc}
\hline
\multirow{2}{*}{Dataset} & \multicolumn{2}{c}{Pre-training}  &  & \multicolumn{2}{c}{Fine-tuning} &  & \multicolumn{1}{c}{\multirow{2}{*}{Test}} \\ \cline{2-3} \cline{5-6}
                         & \multicolumn{1}{c}{Train} & Validation &  & \multicolumn{1}{c}{Train}        & \multicolumn{1}{c}{Validation}          &  & \multicolumn{1}{c}{}                      \\ \cline{1-3} \cline{5-6} \cline{8-8} 
Synthetic                   &   75\%                 & 15\% &  &        -        &   -      &  &     -                                         \\
Same image            &        -                   &  -   &  &   \multicolumn{2}{c}{75\% (k-fold cross validation)}  &  & 25\%                                        \\
Different image           &     -                      &     -  &  &    -      &      -         &  & 100\%                                       \\
\hline
\end{tabular}
\vspace{1mm}
\caption{How to use each dataset in the MLP model experiment. Same image and Different image mean the same image dataset and the different image dataset, respectively. Percentages, such as 75\% or 15\%, represent how much of each dataset is used.}
\label{tab:how-to-use-dataset-mlp}
\vspace{1mm}
\end{table*}
\section{Supplement of Results}
\label{appendix:sup_of_experiment}
Table \ref{tab:learning_results2}, \ref{tab:input_exp2} show F-score, Recall, and Precision in the MLP model's experiment. 
Table \ref{tab:results_openflamingo} shows the results of OpenFlamingo's experiment.
\begin{table*}[h]
\centering
\begin{tabular}{clccllcc}
\hline
                                       & \multicolumn{3}{c}{Same image}                                                                                              &  & \multicolumn{3}{c}{Different image}                                                                                            \\ \cline{2-4} \cline{6-8} 
\multirow{-2}{*}{Learning Method}  & F-score                               & Recall                                & Precision                             &  & F-score                               & Recall                                & Precision                             \\ \cline{1-4} \cline{6-8} 
Human mistake                                    & 0.730 & \textbf{0.920}                        & 0.605 &  & 0.368 & 0.459 & 0.308 \\
Synthetic + Human mistake                                & \textbf{0.811}                        & 0.860 & \textbf{0.768}                        &  & \textbf{0.482}                        & \textbf{0.541}                        & \textbf{0.434}                        \\ \hline
\end{tabular}
\vspace{1mm}
\caption{The 
detailed results of each learning method in the experiment about pretraining with Synthetic Dataset. The score
is F-score, Recall, and Precision.}
\label{tab:learning_results2}
\vspace{1mm}
\end{table*}
\begin{table*}[h]
\centering
\begin{tabular}{cccccccc}
\hline
\multicolumn{1}{c}{\multirow{2}{*}{Model}} & \multicolumn{3}{c}{Same image} &  & \multicolumn{3}{c}{Different image} \\ \cline{2-4} \cline{6-8} 
\multicolumn{1}{c}{}                       & F-score  & Recall  & Precision &  & F-score    & Recall   & Precision   \\ \cline{1-4} \cline{6-8} 
Baseline         & \textbf{0.811} &  \textbf{0.860} & \textbf{0.768} &  &  0.482 & 0.541 & 0.434 \\
QA turn        & 0.718  & 0.840  &  0.627  &  & {\color[HTML] {3531FF}\textbf{0.514}}  & {\color[HTML] {3531FF}\textbf{0.623}} &  {\color[HTML] {3531FF}\textbf{0.437}}    \\
Question type    &  {\color[HTML] {3531FF}\textbf{0.743}} & 0.840 & {\color[HTML] {3531FF}\textbf{0.667}}  &  &  \textbf{0.527} &  \textbf{0.639}  & \textbf{0.448}   \\ \hline
\end{tabular}
\vspace{1mm}
\caption{The detailed results in the MLP model experiment. The score is F-score, Recall, and Precision. The best score is in \textbf{black bold}, and the second-best score is in {\color[HTML] {3531FF}\textbf{blue}}.}
\label{tab:input_exp2}
\end{table*}
\begin{table*}[h]
\centering
\begin{tabular}{@{}cccccccc@{}}
\toprule
\multirow{2}{*}{Prompt type} & \multicolumn{3}{c}{Without history} &  & \multicolumn{3}{c}{With history} \\ \cmidrule(l){2-8} 
                      & F-score    & Recall   & Precision   &  & F-score   & Recall   & Precison  \\ \cmidrule(r){1-4} \cmidrule(l){6-8} 
Normal                &   0.313  &   0.350 & 0.283      & &  0.325 & 0.438  & 0.259         \\
QA turn hint  & \textbf{0.374} & \textbf{0.463} & {\color[HTML] {3531FF}\textbf{0.314}} &  &  \textbf{0.377}  & \textbf{0.538}  &  {\color[HTML] {3531FF}\textbf{0.291}}       \\
Question type hint   &  {\color[HTML] {3531FF}\textbf{0.366}}     & {\color[HTML] {3531FF}\textbf{0.438}}   &  \textbf{0.315}  &  & {\color[HTML] {3531FF}\textbf{0.372}} &  {\color[HTML] {3531FF}\textbf{0.500}} &  \textbf{0.296} \\ \bottomrule
\end{tabular}
\vspace{1mm}
\caption{The detailed results in the OpenFlamingo experiment. The score is F-score, Recall, and Precision. The best score is in \textbf{bold black}, and the second-best score is in {\color[HTML] {3531FF}\textbf{blue}}.}
\label{tab:results_openflamingo}
\end{table*}
\clearpage
{\small
\bibliographystyle{ieee_fullname}
\bibliography{egbib}
}
\end{document}